\documentclass[
twocolumn
]{ceurart}

\usepackage{listings}
\usepackage{graphicx}
\usepackage{subcaption}
\usepackage{tikz}
\usetikzlibrary{arrows.meta, positioning}
\begin{document}

\copyrightyear{2025}
\copyrightclause{Copyright for this paper by its authors.
  Use permitted under Creative Commons License Attribution 4.0
  International (CC BY 4.0).}

\conference{Proceedings of the Seventh Workshop on Automated Semantic Analysis of Information in Legal Text (ASAIL 2025), June 16, 2025, Chicago, USA.}

\title{Incorporating Legal Structure in Retrieval-Augmented Generation: A Case Study on Copyright Fair Use}

\author[1]{Justin Ho}[%
orcid=0009-0005-0751-9504,
email=jho@g.harvard.edu,
url=https://justinhjy1004.github.io/,
]
\cormark[1]
\address[1]{Harvard Business School, Boston MA, 02163, United States of America}

\author[2]{Alexandra Colby}[%
email=acolby@jd26.law.harvard.edu,
]

\author[2]{William Fisher}[%
email=tfisher@law.harvard.edu,
url=https://hls.harvard.edu/faculty/william-w-fisher/
]

\address[2]{Harvard Law School, Boston MA, 02138, United States of America}

\cortext[1]{Corresponding author.}

\begin{abstract}
  This paper presents a domain-specific implementation of Retrieval-Augmented Generation (RAG) tailored to the Fair Use Doctrine in U.S. copyright law. Motivated by the increasing prevalence of DMCA takedowns and the lack of accessible legal support for content creators, we propose a structured approach that combines semantic search with legal knowledge graphs and court citation networks to improve retrieval quality and reasoning reliability. Our prototype models legal precedents at the statutory factor level (e.g., purpose, nature, amount, market effect) and incorporates citation-weighted graph representations to prioritize doctrinally authoritative sources. We use Chain-of-Thought reasoning and interleaved retrieval steps to better emulate legal reasoning. Preliminary testing suggests this method improves doctrinal relevance in the retrieval process, laying groundwork for future evaluation and deployment of LLM-based legal assistance tools.
\end{abstract}


\maketitle

\section{Introduction}

The Digital Millennium Copyright Act (DMCA) provides platforms such as YouTube with safe harbor protection from copyright infringement claims related to content uploaded by their users, as long as they offer copyright holders the ability to take down content that infringes their rights \cite{16_DMCA}.

In theory, an uploader whose content was removed under the DMCA can submit a counter-notice to challenge the takedown, with the most commonly used defense being the Fair Use Doctrine under 17 U.S. Code § 107 \cite{15_USC107}. This doctrine considers four factors: the purpose and character of the use (e.g., whether it is transformative, commercial, and if the work serves a different purpose to the original), the nature of the copyrighted work, the amount and substantiality of the portion used, and the effect of the use on the market for the original or its derivatives. It is designed to permit use of copyrighted material without permission for purposes such as criticism, comment, news reporting, teaching, scholarship, or research.

However, in practice, DMCA takedowns are often abused to suppress valid criticism protected under free speech, or are issued through automated systems that generate invalid and duplicative claims. This results in chilling effects and self-censorship, especially among creators without legal representation, who may be ill-equipped to assess whether they have a colorable fair use defense. Although the courts held in the \textit{Lenz v. Universal Music Corp.}, 801 F.3d 1126 (9th Cir. 2015), decision that copyright holders must consider fair use in good faith before issuing a takedown notice, enforcement of this standard is weak. Users must prove the copyright holder acted in bad faith, a subjective mental state that is difficult to determine, rendering the safeguard largely ineffective in practice \cite{21a_DMCAAbuse, 21b_DMCAAbuse}.

\subsection{LLMs in Legal Assistance}

With the advancement of Large Language Models (LLMs), particularly in the legal domain, there is growing potential for these technologies to offer legal assistance to content creators who might otherwise lack representation to assert fair use claims \cite{10_LLMFewShotLearner, 20_LLMSurvey, 05_BuildingJusticeBot}. While LLMs can automate tasks such as annotation, issue-spotting, interpretation of short legal texts, and even generating legally plausible conclusions, they still fall short in areas that require precise rule recall, multi-step reasoning, and the explanation of legal inferences \cite{06_GPTAnnotateTextualData, 22_LegalBench}. Even Retrieval-Augmented Generation (RAG) models from major legal research platforms are prone to hallucinations, including fabricating case law and misinterpreting precedents \cite{04_LegalHallucination, 04b_HallucinationFree}.

Following the typology of RAG-based hallucinations proposed in \cite{04b_HallucinationFree}, persistent issues arise from a combination of naive retrieval, inapplicable authority, sycophancy (i.e., the tendency to agree with a given text even when it is inaccurate), and reasoning errors. We hypothesize that local domain improvements—specifically, building expertise within a narrow subfield of legal doctrine—can improve the deployment performance of LLMs in certain legal contexts. Such narrowly focused local subfield experts can potentially be combined to provide more general automated legal assistance. Currently, we focus on the Fair Use Doctrine in copyright law as a case study.  This is conceptually similar to Mixture of Experts (MoE) models \cite{13_AdaptiveMoE}. However, our focus is on improving the non-parametric memory component of RAG by combining knowledge graphs and granular retrieval strategies in the Fair Use Doctrine \cite{23_NonParametricRAGContinualLearning, 01_UnifyingKGwithLLM, 03b_SemanticRepresentationContextual, 02_DenseRetrieval}.

\subsection{Local Expertise and Structured Reasoning}

Problems like naive retrieval and inapplicable authority may have stemmed from the general-purpose Question-Answering (QA) design of AI models deployed by major legal research platforms, but due to their proprietary nature, it is difficult to verify the true source of the problems \cite{04b_HallucinationFree}. Legal concepts may appear semantically similar in common usage, yet differ significantly in terms of legal doctrine (e.g., the distinction between `moral turpitude' and the `moral-wrong doctrine' in Criminal Law, or the differing meanings of `negligence' and `reasonable person' across various areas of law). Although our focus on the Fair Use Doctrine offers some topical constraint, we show that retrieval can be improved by incorporating legal information as a citation-weighted knowledge graph. This graph encodes court hierarchy, citation relationships, and the statutory factors specific to Fair Use. As a result, the retrieval process prioritizes documents that are not only semantically relevant but also doctrinally authoritative.

We also include methods used by ``reasoning models," such as Chain-of-Thought (CoT), to improve multi-step reasoning in legal cases since CoT has been shown to reduce reasoning errors in LLMs \cite{08_CoT}. This is especially important for decisions under the Fair Use Doctrine, which is a multi-factor test requiring contextual considerations. Additionally, we implement a one-step Interleaving Retrieval CoT, where the LLM first analyzes how a complaint or case relates to the four fair use factors, which then guides the retrieval process \cite{28_CoTandIRCoT}. This may reduce sycophancy by anchoring the model’s reasoning in the structure of the doctrine itself. However, the issue of sycophancy is perhaps better addressed during the information elicitation stage.

We developed a functioning prototype to demonstrate the core features of our system, which is available at \href{https://fairuselegalbot-main.streamlit.app/}{https://fairuselegalbot-main.streamlit.app/}. The source code of the prototype, construction of the knowledge graph, and the results of the preliminary analysis is publicly available on GitHub: \href{https://github.com/justinhjy1004/FairUseLegalBot}{https://github.com/justinhjy1004/FairUseLegalBot}.

\section{Literature Review and Related Work}
Our work integrates ideas from Retrieval-Augmented Generation (RAG), knowledge graphs, and information retrieval and representation, adapting them to the legal domain.

\subsection{Similarity is Not All You Need}
A widely adopted strategy to mitigate hallucinations in language models is grounding them through Retrieval-Augmented Generation (RAG). RAG retrieves external documents based on vector similarity to the user’s query, typically measured using cosine similarity \cite{09_OriginalRAGPaper}.

In legal applications, the retrieved documents often include case law, legal opinions, statutes, and regulatory codes. The core assumption is that retrieving semantically similar documents will produce more factually accurate and contextually relevant outputs by anchoring them in authoritative sources. However, the quality of the model’s output is only as strong as the documents retrieved. In practice, LLMs deployed within legal research tools still exhibit hallucinations—especially when the retrieval corpus is noisy, outdated, or lacks contextual metadata, such as indicators that a precedent has been overruled \cite{04_LegalHallucination}.

Recent research in information retrieval and domain-specific indexing has aimed to improve the reliability of RAG-based systems. Still, the notion of similarity remains highly nuanced and context-dependent \cite{03b_SemanticRepresentationContextual}. For instance, “Dracula” might refer to either the character or the novel, and a summarization task could be misled by retrieving content about Nosferatu, an unauthorized adaptation, despite the surface-level similarity\footnote{The original Nosferatu (1922) was found by German courts to infringe the Stoker Estate’s copyright. A German judge ordered all copies of Nosferatu to be destroyed and it survives today because of a single copy that found its way to the United States.}.

Additionally, the granularity of the retrieval unit, whether at the document, sentence, or sub-sentence level, is important in downstream retrieval performance. Often, only a small portion of a document is relevant to the query, and this is particularly true in the context of legal reasoning and analysis \cite{02_DenseRetrieval, 32_LegalCitationNetwork}. To address this, we structure our underlying data store at the level of statutory factors, allowing retrieval to operate at a finer granularity aligned with the specific analysis of the Fair Use Doctrine.

\subsection{Incorporating Legal Structure}

Legal reasoning relies on more than surface-level textual similarity. Documents in the legal domain carry contextual structure that flat document representation, as commonly used in standard RAG implementations, often fail to capture. Legal opinions are authored by courts of differing authority, and in common law systems, precedents shape legal interpretation. Determining which precedents are most relevant requires understanding the legal hierarchy, interpretive weight, and the frequency and influence of citation. Representing this information as a knowledge graph, where relationships between cases are explicitly modeled, can improve retrieval quality \cite{07a_KGRAG, 07b_GraphRAG}.

Our approach builds on this idea by encoding legal structure directly by modeling court hierarchies, citation flows, and the interpretive weight of specific paragraphs with respect to statutory factors under consideration. This improves both the doctrinal relevance of retrieved material and the accuracy of subsequent inference tasks.

Prior work in U.S. and EU legal systems demonstrates the value of citation networks, particularly when nodes represent cited paragraphs rather than entire opinions. Prior work shows that paragraph-level modeling captures the `grammar of repetition' in judicial reasoning. This illustrates how interpretive principles gain authority through repeated citation. Such granularity also enables detection of indirect influence chains and improves our understanding of how legal doctrines evolve \cite{32_LegalCitationNetwork}.

In this spirit, we structure our dataset around statutory factor-level modeling. By explicitly annotating legal opinions according to the statutory factors from the Fair Use Doctrine. This enables context-sensitive retrieval that has the potential of improving performance. For instance, two copyright disputes involving unauthorized film use might seem similar, but diverge sharply depending on whether the use is non-expressive or a parody of the original material. This distinction is important in fair use analysis, and modeling the data in this granularity helps reflect and align how courts often extract legal principles from specific parts of a ruling rather than relying on the full opinion \cite{32_LegalCitationNetwork}.

We also incorporate the citation network structure of legal precedents by modeling court hierarchies and citation relationships to include not only the semantic similarity of the dispute at hand, but also the doctrinal relevance and importance. We apply PageRank \cite{page1999pagerank}, commonly used in citation analysis, as a way to incorporate the doctrinal relevance in our retrieval and ranking process. While basic degree centrality offers insight, legal reasoning can drift or `un-anchor' from original sources over time \cite{32_LegalCitationNetwork}. PageRank, by contrast, accounts for the authority of citing sources—i.e., a citation from a widely cited opinion carries more doctrinal weight than one from a marginal case \cite{page1999pagerank}. This allows our system to better reflect the practical significance of legal authority in ranking the retrieved judicial opinions.

\section{Methods}
\label{sec: Methods}

This section provides the details regarding the implementation of the automated analysis of Fair Use cases, particularly in data representation and the retrieval process.

\begin{figure}[h]
    \centering
    \includegraphics[width=\linewidth]{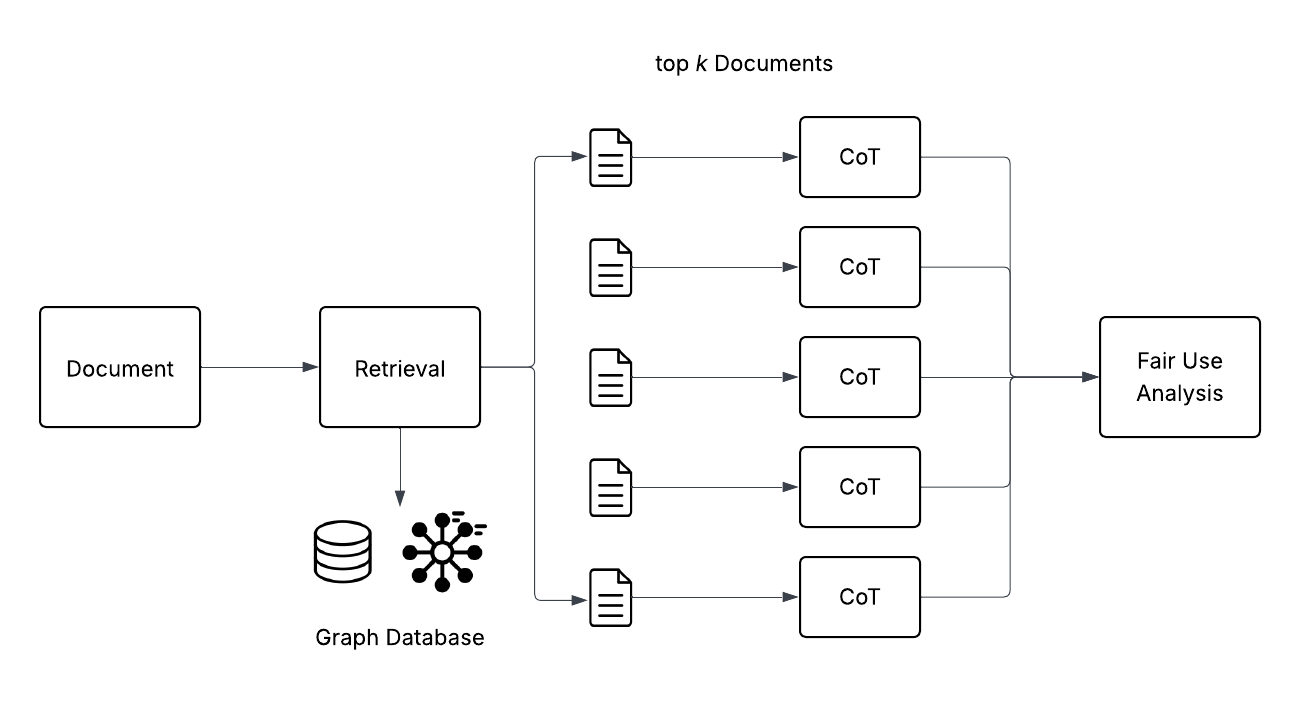}
    \caption{Overview of the Automated Analysis of Fair Use Cases.}
    \label{fig:fair_use_legal_bot}
\end{figure}

Figure \ref{fig:fair_use_legal_bot} shows an overview
of the methods used. The system starts by retrieving the top-$k$ most relevant legal cases using a graph-based database that takes into account not just text similarity, but also court authority and how often cases are cited. Each case is then analyzed step by step using Chain-of-Thought reasoning to assess how it applies to the four Fair Use factors. Finally, these analyses are combined to generate a structured Fair Use evaluation based on the user’s document.

The Large Language Model used is Google's Gemini Flash 2.0 \cite{11_GeminiLLM}, and the embedding model for semantic-based vector search is Google's Gecko \cite{29_GeckoEmbeddings}.

\subsection{Data Corpus}

Using WestLaw Precision's fact pattern search, we located all legal precedents relevant to the Fair Use Doctrine in copyright law. We then sourced the legal corpus relevant to the Fair Use Doctrine in copyright from Court Listener \cite{25_free_law_project_recap_2020} and Hein Online. Furthermore, we used EyeCite, an Open Source Software developed to identify case law citations in documents to construct our citation network \cite{24_eyecite}. 

\begin{table}[h!]
  \centering
  \caption{Corpus Overview}
  \begin{tabular}{lc}
    \toprule
    Total Number of Cases & 209 \\
    Total Number of Opinions & 283 \\
    Time Range Coverage & 1976-2025\\
    Number of Unique Courts & 51 \\
    \bottomrule
  \end{tabular}
\end{table}

The number of opinions exceeds the number of cases because a single case may generate multiple judicial opinions, including appellate decisions, as well as concurring or dissenting views authored by individual judges.

Additionally, we sourced complaints related to Copyright infringement that were not resolved in court from Public Access to Court Electronic Records (PACER) \cite{26_PACER}. These serve as a preliminary test dataset for our working prototype, as they represent real Fair Use disputes that were unresolved, and hence provide a way to measure how well our model performs in unresolved cases. We sourced a total of 20 cases.

\subsection{Data Representation}
\label{sec: data_representation_KG}

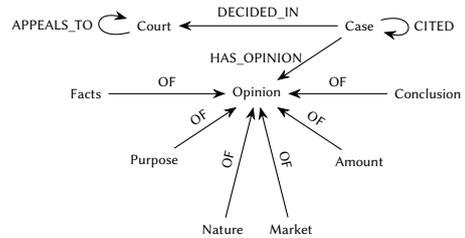
\begin{figure}[ht]
\centering
\begin{tikzpicture}[
    ->, >=Stealth,
    nodeStyle/.style={
        align=center,
        font=\scriptsize,
        fill=white
    },
    labelStyle/.style={
        font=\scriptsize
    },
    scale=0.9,
    transform shape
]

\node[nodeStyle] (courts)      at (1,-1)   {Court};
\node[nodeStyle] (cases)       at (4,-1)   {Case};

\node[nodeStyle] (opinion)     at (2.5,-2)  {Opinion};

\node[nodeStyle] (facts)       at (0,-2)  {Facts};
\node[nodeStyle] (purpose)     at (1,-3) {Purpose};
\node[nodeStyle] (nature)      at (2,-4)  {Nature};
\node[nodeStyle] (market)      at (3,-4) {Market};
\node[nodeStyle] (amount)      at (4,-3)  {Amount};
\node[nodeStyle] (conclusion)  at (5,-2)   {Conclusion};

\draw[->]
  (courts) edge [loop left] node[labelStyle] {APPEALS\_TO} (courts);

\draw[<-] 
  (courts) -- node[labelStyle, above] {DECIDED\_IN} (cases);

\draw[->]
  (cases) edge [loop right] node[labelStyle] {CITED} (cases);

\draw[->]
  (cases) -- node[labelStyle, left] {HAS\_OPINION} (opinion);

\draw[->]
  (facts) -- node[labelStyle, above, sloped] {OF} (opinion);

\draw[->]
  (purpose) -- node[labelStyle, above, sloped] {OF} (opinion);

\draw[->]
  (nature) -- node[labelStyle, above, sloped] {OF} (opinion);

\draw[->]
  (market) -- node[labelStyle, above, sloped] {OF} (opinion);

\draw[->]
  (amount) -- node[labelStyle, above, sloped] {OF} (opinion);

\draw[->]
  (conclusion) -- node[labelStyle, above, sloped] {OF} (opinion);
\end{tikzpicture}
\caption{Domain Specific Knowledge Graph Schema of Judicial Opinions of the Fair Use Doctrine in Copyright 17 U.S. Code § 107.}
\label{fig:KG_Schema}
\end{figure}

Our knowledge graph is implemented using Neo4j \cite{27_neo4j}. In order to more faithfully represent the data in legal precedents related to the Fair Use Doctrine in copyright, we modeled the cases with a knowledge graph, and the schema of the graph database is shown in Figure \ref{fig:KG_Schema}. A schema defines the structure of the graph—it specifies what types of entities (nodes) exist, such as \textit{Case}, \textit{Court}, \textit{Opinion}, and Fair Use Factor (e.g., \textit{Purpose}, \textit{Market}), and how they are connected through relationships like \textit{CITED}, \textit{DECIDED\_IN}, or \textit{HAS\_OPINION}. For example, a \textit{Case} node from the Supreme Court is connected to a \textit{Court} node labeled ``SCOTUS'' and is cited by multiple lower-court \textit{Opinion} nodes.  This incorporates important features of the legal system in the United States where legal precedents issued by a higher court have higher authority over others, as well as the citation network formed between the cases.

Furthermore, in every given opinion, we extract the verbatim paragraphs related to the \textit{Facts} of the case, the four factors: (1) \textit{Purpose} and character of the use (2) \textit{Nature} of the copyrighted work, (3) \textit{Amount} and substantiality of the portion used, and (4) Effect of the use on the potential \textit{Market}. We also included the \textit{Conclusion} of the opinion to reflect how the court balanced the four factors to arrive at their opinion. The extraction is done using the LLM to identify and extract verbatim paragraphs in which each of the four Fair Use factors were discussed, along with the factual background and the court's conclusion. We designed specific prompts to direct the LLM to focus on legal reasoning and factor-specific content. These prompts instruct the model to return direct quotations from the opinion text corresponding to each factor, rather than paraphrased summaries. 

The choice of our data representation is that it is not only a more faithful representation of the data, which allows us to use the structure of the data (i.e. citations) to improve our retrieval process, but it also provides the ability to retrieve based on contextual similarity. For instance, a complaint on copyright infringement might be similar with respect to the medium in which the work was distributed (print or via video recordings), but might differ substantially based on the purpose of the use (ie. parody, criticism, or for educational reasons). 

The choice of using a knowledge graph representation allows for more granular, context-specific similarity comparison during the retrieval process which has shown to be effective \cite{02_DenseRetrieval, 03b_SemanticRepresentationContextual}. Moreover, the interpretability of such representation might increase the interest, trust, and therefore adoption of LLMs in the legal space since this mimics how legal experts might reason in the context of Fair Use \cite{18_TrustAIExplainability}.

Each case is modeled as a node connected to its issuing court and to the legal opinion(s) it contains. Opinions are linked to factor-specific paragraph nodes (e.g., Purpose, Market, Nature), enabling granular retrieval by legal reasoning dimensions. Citations between cases form directed edges within the graph. The schema is implemented in Neo4j using labeled nodes (e.g., \texttt{Case}, \texttt{Court}, \texttt{Opinion}, \texttt{Fact}) and relationship types (e.g., \texttt{DECIDED\_IN}, \texttt{HAS\_OPINION}, \texttt{CITED}, \texttt{APPEALS\_TO}). This representation supports both structural queries (e.g., retrieving appellate court opinions) and vector search via LLM embeddings.

\subsection{Retrieval and Reranking}

Our retrieval process combines semantic-based vector search by computing similarity scores. However, we extend this by incorporating two features from the data representation by determining the authoritativeness of a legal precedent using the PageRank algorithm as well as the cited opinions of the opinions that were retrieved.

As discussed in Section \ref{sec: data_representation_KG}, the verbatim passages that discuss the facts of the case, the four factors of Fair Use, and the conclusion of the case were extracted using an LLM. We then chunk the passages and embed them using Gecko. We use cosine similarity as our method in computing the similarity between the documents. Furthermore, to incorporate the citation metrics as well as the court hierarchy, we used the PageRank algorithm to quantify the relative importance of each court decision within the legal citation network. This is done for both the legal opinions (based on the citations) and the courts (based on appellate relationships). 

We used the PageRank algorithm to quantify two distinct but complementary aspects of legal authority: citation authority, calculated from the inter-opinion citation network, and court hierarchy, based on appellate relationships among courts. While these dimensions capture different sources of legal relevance—influence through citation versus institutional authority by position in the judiciary, they are often correlated in practice, as higher courts tend to issue opinions that are more frequently cited. However, we include this dual representation to allow our model to consider both the structural and reputational weight of each legal source.

The retrieved documents are ranked based on a convex combination, where
\begin{equation}
    s_{i} = w_{\text{text}}\text{TextSim}_i + w_{\text{cit}}\text{Citation}_i + w_{\text{court}}\text{Court}_i
\end{equation}
such that $w_{\text{text}}, w_{\text{cit}}, w_{\text{court}} \in [0,1]$ and $w_{\text{text}} + w_{\text{cit}} + w_{\text{court}} = 1$. The weights hence can be interpreted as hyperparameters in which one can adjust for optimal retrieval. We applied \texttt{min-max} scaling to the scores individually to ensure that each score is between 0 and 1. In the current prototype, the weights are manually specified by the user, which allows legal experts to adjust the retrieval behavior based on the characteristics of the query or dispute. For instance, a legal expert may prioritize citations and court hierarchy in appellate-heavy disputes, while another may favor textual similarity in novel or atypical cases. In future work, we plan to systematically evaluate the effect of different weight configurations using ablation studies.

Lastly, based on the top $k$ legal precedents that are retrieved, an additional parameter $n$ can be specified to retrieve the cited opinions by the retrieved cases based on the citation and court rankings. These cited cases are included directly in the inference step to provide broader legal context and to simulate how a legal practitioner might draw from precedent when reasoning about a novel dispute.

\begin{figure}[h]
    \centering
    \includegraphics[width=\linewidth]{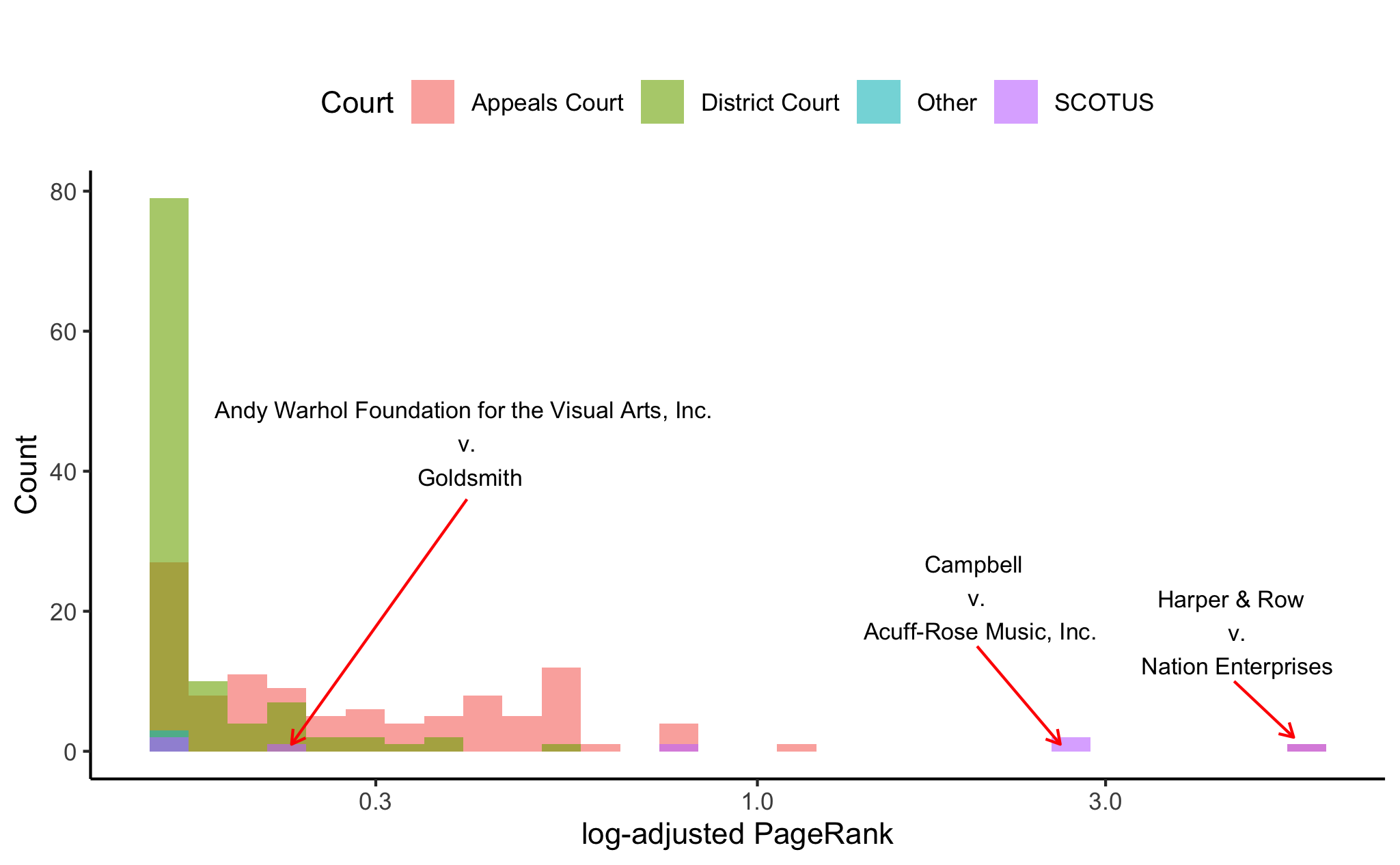}
    \caption{Distribution of Legal Case Influence by Court and PageRank}
    \label{fig:pagerank}
\end{figure}

Figure \ref{fig:pagerank} displays the distribution of legal cases by their log-adjusted PageRank, a measure of influence within the legal citation network. Most cases, particularly from District Courts, cluster at lower PageRank values, while a few landmark Supreme Court decisions like \textit{Campbell v. Acuff-Rose Music, Inc.} and \textit{Harper \& Row v. Nation Enterprise} have disproportionally high PageRank values. This is, of course, not surprising, as many natural networks exhibit power-law distributions \cite{33_ScaleFreeNetwork}.

Although \textit{Warhol v. Goldsmith}, 598 U.S. 508 (2023), is considered to be highly significant by most legal scholars, its recency means it has had limited time to accumulate citations. This is a limitation of PageRank which does not account for time. Future work could explore time-adjusted measures to better capture the emerging influence of newer cases.

\subsection{Current Progress and Implementation}

\begin{figure}[h]
    \centering
    \includegraphics[width=\linewidth]{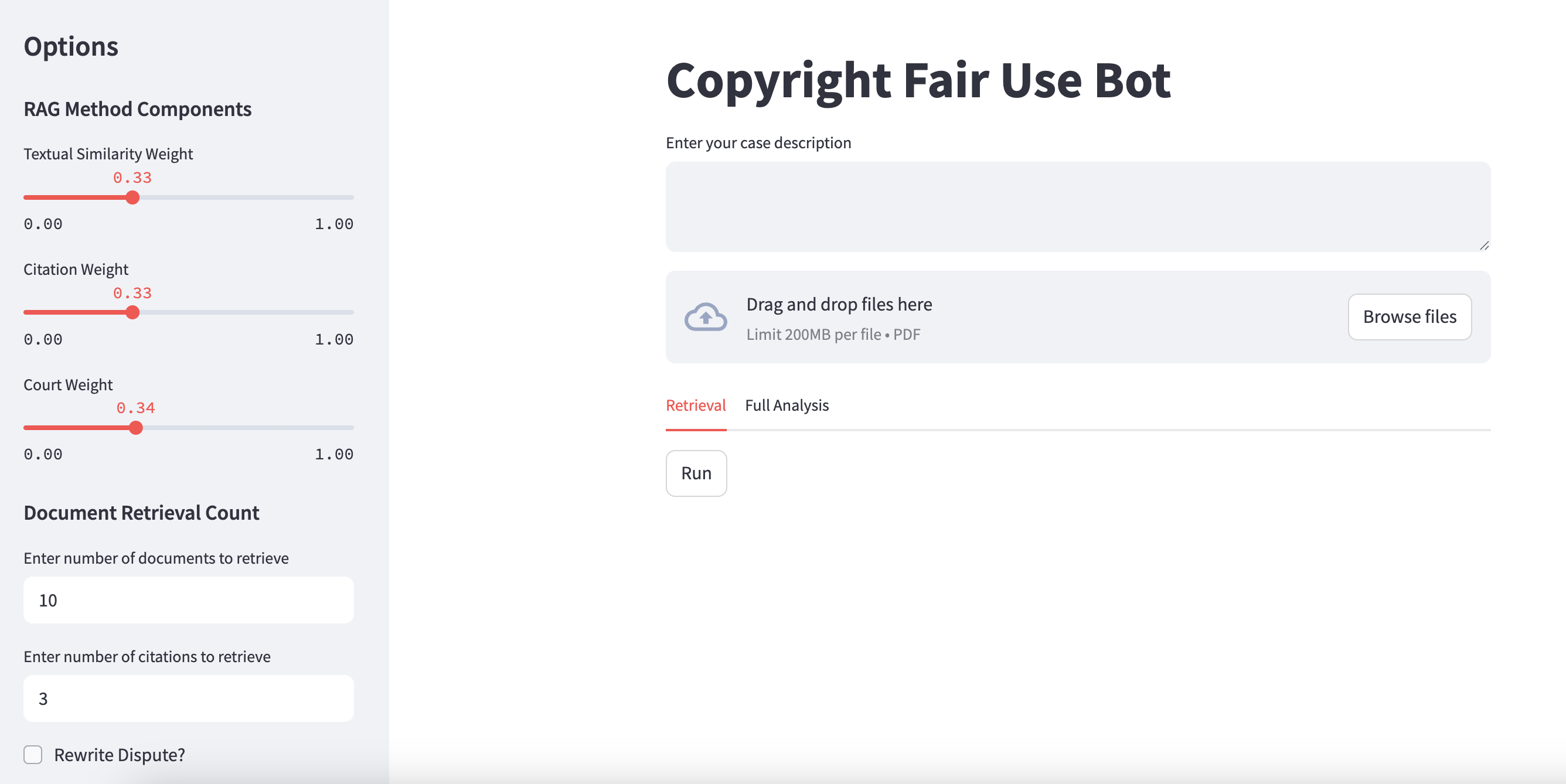}
    \caption{Interface of the Fair Use Legal Bot}
    \label{fig:interface}
\end{figure}

As of April 2025, we have developed a functional prototype of the application. The current prototype of the Fair Use Legal Bot can be found \href{https://fairuselegalbot-main.streamlit.app/}{here}. Figure \ref{fig:interface} shows the interface where users can provide their case or dispute description for fair use analysis. Users can upload relevant documents in PDF format or enter a written description directly into the input box. They can also customize the retrieval algorithm by adjusting the weights for textual similarity, citation frequency, and court relevance, as well as specify the number of documents and citations to retrieve. This prototype was developed mainly for internal testing and refinement of the retrieval process, but is accessible to external users for testing and evaluation.

The current version supports uploading a complaint or a text description of a dispute, and retrieve the most relevant documents based on the hyperparameters configured in the left panel (``RAG Component Methods”). We have currently implemented the manual weighting of the three parameters, and users can specify the $k$ number of documents as well as $n$ number of cited cases. 

While formal evaluations and user studies have not yet been conducted, the current version of the application establishes a strong foundation for future experimentation and ablation studies.

\section{Preliminary Testing on Retrieval}

In this section, we describe the preliminary experiments conducted to evaluate our retrieval system. The primary goal was to compare the baseline Standard RAG approach with our proposed Structured RAG method, which incorporates additional legal structure in the form of citation authority and court hierarchy.

\begin{figure*}[htbp]
  \centering
  \begin{subfigure}[b]{0.52\textwidth}
    \centering
    \includegraphics[width=\textwidth]{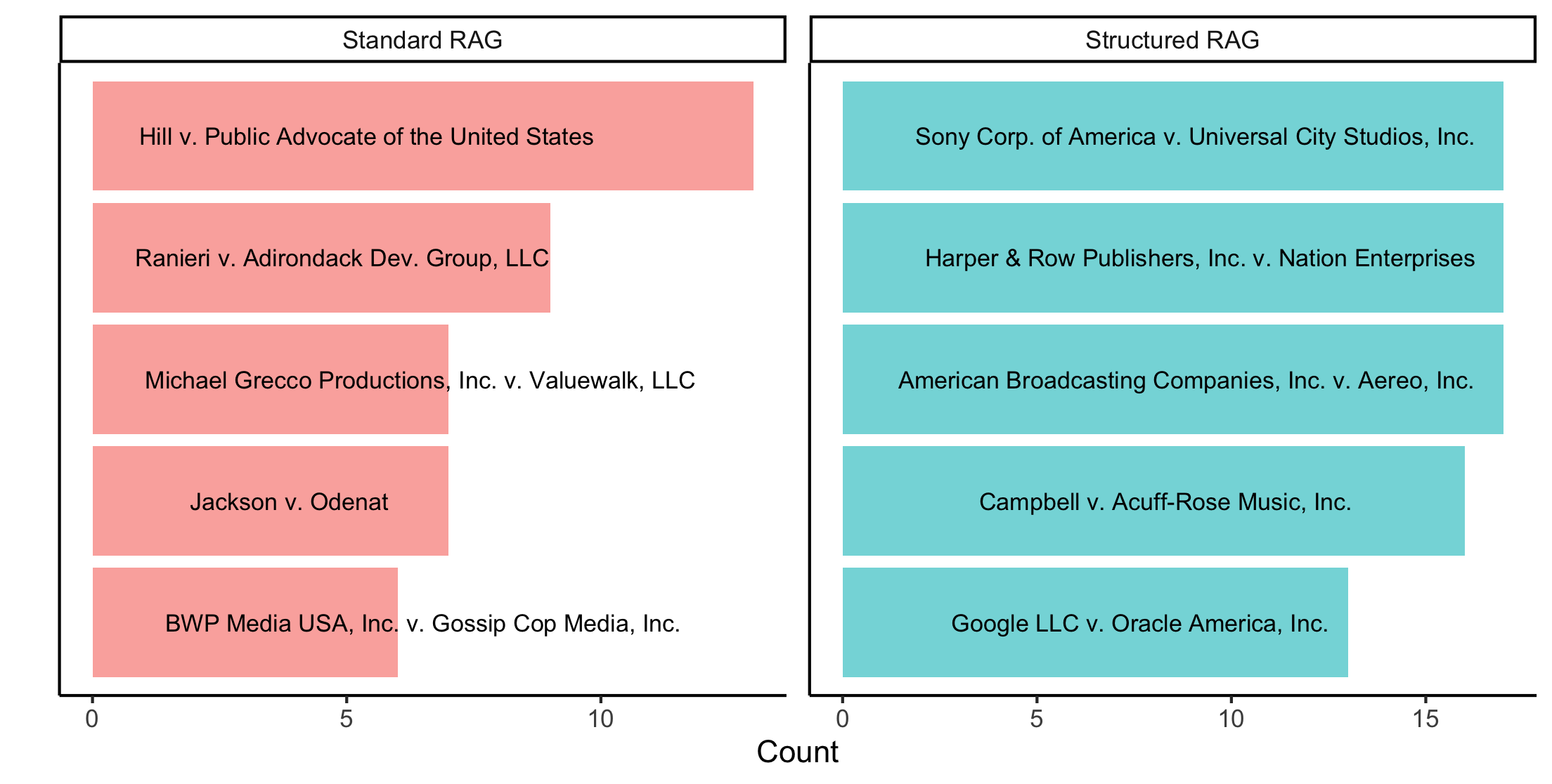}
    \caption{Most Commonly Retrieved Cases in Each Retrieval Method}
    \label{fig:topcases}
  \end{subfigure}
  \hfill 
  \raisebox{3ex}{
  \begin{subfigure}[b]{0.4\textwidth}
    \centering
    \includegraphics[width=\textwidth]{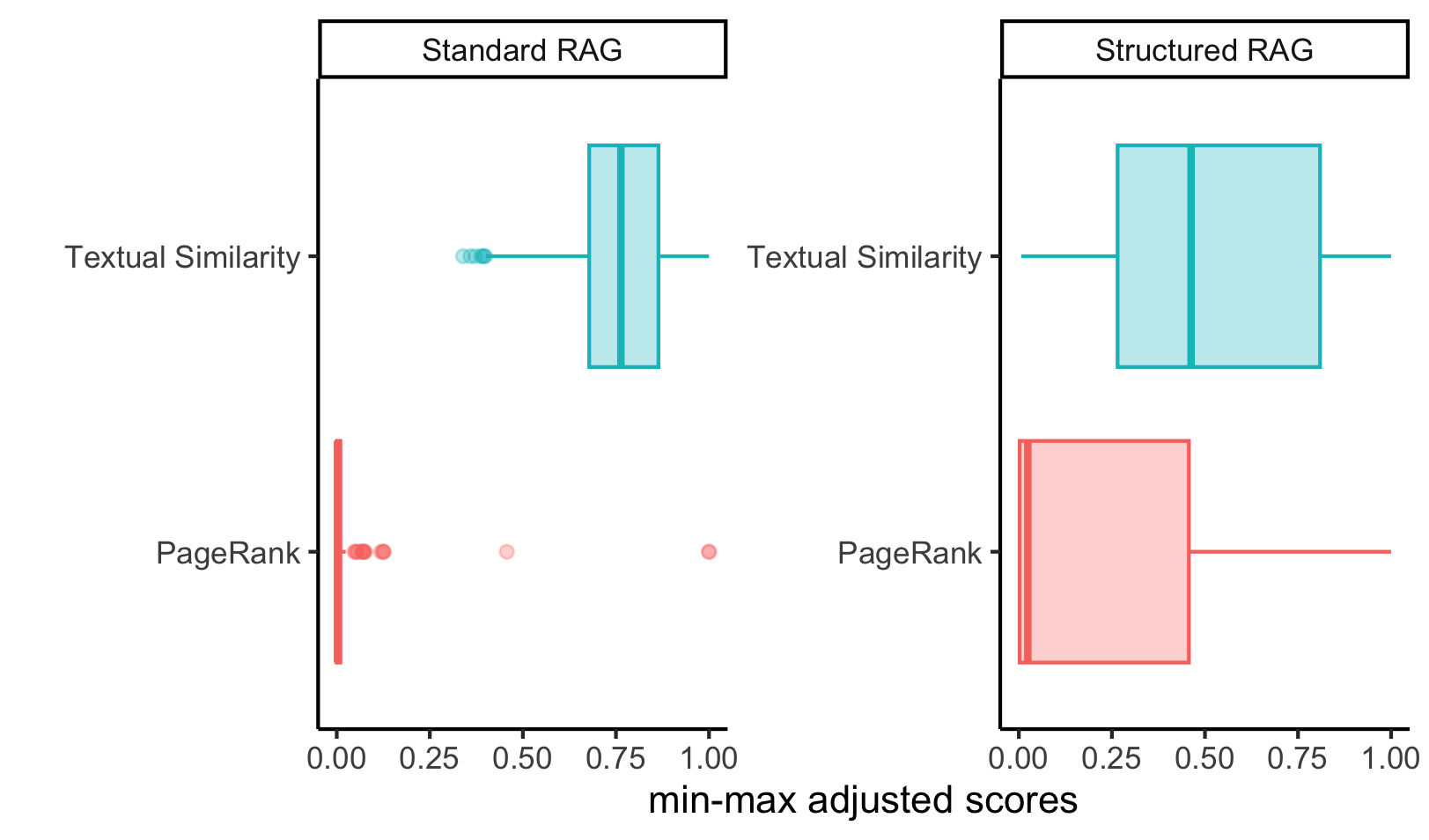}
    \caption{Boxplot of scores for retrieved cases under Standard RAG and Structured RAG.}
    \label{fig:metric}
  \end{subfigure}
  }
  \caption{Comparison of Retrieval Methods. The Standard RAG achieves higher textual similarity but lacks doctrinal authority as measured by PageRank, and the most common cases retrieved tend to be less authoritative.}
  \label{fig:combined}
\end{figure*}

\begin{table}[ht]
  \centering
  \caption{Comparison of Retrieval Methods with Grouped Metrics. All scores are min-max scaled.}
  \label{tab:grouped_retrieval_comparison}
  \begin{tabular}{l cc | cc}
  \toprule
  \textbf{Retrieval Method} & \multicolumn{2}{c}{\textbf{PageRank}} & \multicolumn{2}{c}{\textbf{Text Similarity}} \\
  \cmidrule(r){2-3} \cmidrule(l){4-5}
                            & \textbf{Mean}    & \textbf{SD}    & \textbf{Mean}      & \textbf{SD}      \\
  \midrule
  Standard RAG    & 0.026 & 0.114 & 0.753 & 0.169 \\
  Structured RAG  & 0.213 & 0.315 & 0.521 & 0.305 \\
  \bottomrule
  \end{tabular}
  \end{table}

We used the unresolved copyright complaints from PACER for our preliminary testing and the experiments were run under two configurations:

\begin{itemize}
  \item \textbf{Standard RAG:} The retrieval process relies solely on textual similarity. The hyperparameters are set \(w_{\text{text}}\) = 1 and \(w_{\text{cit}} = w_{\text{court}}= 0\)
  \item \textbf{Structured RAG:} The retrieval incorporates legal structural elements by assigning uniform weights to each component \(w_{\text{text}} = w_{\text{cit}} = w_{\text{court}} = 0.333\).
\end{itemize}
  
We compared the PageRank score and cosine similarity score (reflecting textual similarity) of the retrieved legal opinions. Unsurprisingly, the Standard RAG achieves high textual similarity (Mean = 0.753, SD = 0.169) but retrieves cases with low doctrinal authority as reflected by its low PageRank scores (Mean = 0.026, SD = 0.114). In contrast, the Structured RAG yields higher doctrinal relevance with significantly increased PageRank scores (Mean = 0.213, SD = 0.315), though its textual similarity is somewhat lower (Mean = 0.521, SD = 0.305) - Figure \ref{tab:grouped_retrieval_comparison}. These findings support our hypothesis that adding legal structural data enhances the retrieval of legally significant cases, and could be a way to reduce problems arising from naive retrieval and inapplicable authority \cite{04b_HallucinationFree}.

However, we want to note that this preliminary testing is limited since (1) PageRank is an imperfect estimate of the doctrinal authority as noted in Section \ref{sec: Methods} and (2) the tradeoff between textual similarity and doctrinal relevance might lead to worse Fair Use analysis.

\section{Limitations and Future Work}

While our prototype demonstrates promising initial results, there are several limitations that must be addressed in future work to ensure robust and reliable deployment.

\subsection{Future Evaluation}

Our current evaluation is limited to internal testing using unresolved copyright complaints and a curated set of legal precedents. To rigorously assess the effectiveness of our prototype, future work should include user studies with legal practitioners and creators, as well as quantitative metrics such as retrieval precision, argument validity, citation relevance, and user trust. We also plan to perform ablation studies to evaluate the individual contributions of textual similarity, citation authority, and court hierarchy in the retrieval scoring function, along with more granular retrieval methods. Informal, preliminary testing has been promising.

Additionally, it is important to evaluate not only the factual and doctrinal accuracy of generated analyses, but also the quality and persuasiveness of the legal arguments. Since legal reasoning involves a degree of subjectivity and contextual nuance, human-in-the-loop evaluations will be essential for understanding the viability of the prototype.

\subsection{Limitations of Current Work}

Despite our focus on grounding retrieval in legal structure, our prototype still exhibits known weaknesses of LLMs, including hallucination and sycophancy. For instance, when presented with vague or generic inputs, the model may generate speculative or overly confident legal conclusions. This is especially problematic in scenarios where users are not legally trained and may rely too heavily on the prototype’s output without independent verification.

While the prototype is designed with legal structure in mind, its interface and guidance mechanisms are not yet optimized for lay users. Since the goal is to support individuals subjected to unfair DMCA takedowns, there is a need for an appropriate information elicitation phase—where an LLM prompts users to describe their dispute, provide specific details relevant to a Fair Use defense, and potentially disclose points that might disqualify them from Fair Use protection.

Furthermore, as noted in Section \ref{sec: Methods}, the use of PageRank—which has a bias against recency—may result in the omission of relevant judicial opinions that reflect evolving doctrine. Empirical legal work that studies how courts interpret and apply legal doctrines can be integrated into the model to complement the limitations of citation-based metrics by capturing nuanced shifts in judicial reasoning and doctrinal emphasis \cite{17_FairUse}.

\subsection{Extension to Other Legal Doctrines}

The current prototype assumes the input case pertains to Fair Use and does not include functionality for classifying the applicability of legal doctrines. Expanding the prototype to determine whether Fair Use is even the appropriate legal framework for a given dispute remains an important next step. The choice to use knowledge graphs was made with the intent of enabling future integration of other legal doctrines.

Future work can build on and extend the current prototype by constructing modules of local expertise that integrate into a larger system. This will likely require a routing mechanism—for instance, training a classifier to determine which legal doctrine applies to a case, and then routing it to the relevant `expert’.
\section{Discussion and Conclusion}

This paper introduces a structured approach to Retrieval-Augmented Generation (RAG) for legal analysis, using the Fair Use Doctrine in copyright law as a case study. By incorporating knowledge graphs that model citation networks, court hierarchies, and statutory factor-level reasoning, our system aims to address persistent issues in legal LLM applications—namely hallucination, irrelevant retrieval, and inadequate legal inference.

Our method aligns with how legal professionals approach multi-factor tests, providing a more interpretable and granular framework that improves both retrieval and downstream reasoning. The integration of citation-based authority metrics and Chain-of-Thought reasoning supports more grounded and nuanced analysis than traditional vector-based approaches alone.

While our prototype remains in an early stage, the foundational design lays the groundwork for both academic study and practical applications. Future work will focus on empirical validation, interface development for non-expert users, and potential generalization to other areas of law. We believe that structuring AI systems around legal doctrines and reasoning patterns holds significant promise for improving access to justice and legal assistance.

\begin{acknowledgments}
  We would like to thank the Berkman Klein Center for Internet \& Society at Harvard University for their support of this research.
\end{acknowledgments}

\section*{Declaration on Generative AI}
  
 \noindent{\em{During the preparation of this work, the author(s) used ChatGPT in order to: Grammar and spelling check, Paraphrase and reword, Generate literature review, and Improve writing style. After using this tool/service, the author(s) reviewed and edited the content as needed and take(s) full responsibility for the publication’s content.}} 

\bibliography{citation}

\end{document}